\PassOptionsToPackage{unicode}{hyperref}
\PassOptionsToPackage{hyphens}{url}
\PassOptionsToPackage{dvipsnames,svgnames,x11names}{xcolor}
\documentclass[
  english,
  11pt,
]{article}
\usepackage{xcolor}
\usepackage[margin=0.8in]{geometry}
\usepackage{amsmath,amssymb}
\usepackage{iftex}
\ifPDFTeX
  \usepackage[T1]{fontenc}
  \usepackage[utf8]{inputenc}
  \usepackage{textcomp} 
\else 
  \usepackage{unicode-math} 
  \defaultfontfeatures{Scale=MatchLowercase}
  \defaultfontfeatures[\rmfamily]{Ligatures=TeX,Scale=1}
\fi
\usepackage{lmodern}
\ifPDFTeX\else
\fi
\IfFileExists{upquote.sty}{\usepackage{upquote}}{}
\IfFileExists{microtype.sty}{
  \usepackage[]{microtype}
  \UseMicrotypeSet[protrusion]{basicmath} 
}{}
\makeatletter
\@ifundefined{KOMAClassName}{
  \IfFileExists{parskip.sty}{%
    \usepackage{parskip}
  }{
    \setlength{\parindent}{0pt}
    \setlength{\parskip}{6pt plus 2pt minus 1pt}}
}{
  \KOMAoptions{parskip=half}}
\makeatother
\usepackage{graphicx}
\makeatletter
\newsavebox\pandoc@box
\newcommand*\pandocbounded[1]{
  \sbox\pandoc@box{#1}%
  \Gscale@div\@tempa{\textheight}{\dimexpr\ht\pandoc@box+\dp\pandoc@box\relax}%
  \Gscale@div\@tempb{\linewidth}{\wd\pandoc@box}%
  \ifdim\@tempb\p@<\@tempa\p@\let\@tempa\@tempb\fi
  \ifdim\@tempa\p@<\p@\scalebox{\@tempa}{\usebox\pandoc@box}%
  \else\usebox{\pandoc@box}%
  \fi%
}
\def\fps@figure{htbp}
\makeatother
\ifLuaTeX
\usepackage[bidi=basic,shorthands=off]{babel}
\else
\usepackage[bidi=default,shorthands=off]{babel}
\fi
\ifLuaTeX
  \usepackage{selnolig} 
\fi
\providecommand{\tightlist}{%
  \setlength{\itemsep}{0pt}\setlength{\parskip}{0pt}}
\usepackage[numbers,sort,compress]{natbib}
\usepackage{xurl}
\usepackage{caption}
\usepackage{needspace}
\usepackage{booktabs}
\usepackage{array}
\newsavebox{\papertable}
\definecolor{PaperTeal}{HTML}{136C60}
\definecolor{PaperBlue}{HTML}{355D92}
\definecolor{PaperMuted}{HTML}{566168}
\AtBeginDocument{\hypersetup{linkcolor=PaperBlue,citecolor=PaperBlue,urlcolor=PaperBlue}}
\AtBeginDocument{\setlength{\bibsep}{3pt}}
\usepackage{bookmark}
\IfFileExists{xurl.sty}{\usepackage{xurl}}{} 
\makeatletter
\@ifundefined{xmpquote}{}{}
\makeatother
\hypersetup{
  pdftitle={The Control-Flow Tax: Specialized Decision Routing for LLM Agent Workflows},
  pdflang={en},
  colorlinks=true,
  linkcolor={Maroon},
  filecolor={Maroon},
  citecolor={Blue},
  urlcolor={Blue},
  pdfcreator={LaTeX via pandoc}}

\title{The Control-Flow Tax: Specialized Decision Routing for LLM Agent
Workflows}
\author{Zhikai Cheng\qquad Shuanghu Wang\thanks{Shuanghu Wang and Li Jiang share second authorship.}\qquad Li Jiang\footnotemark[1]\\[4pt]Talkdesk}
\date{}

\begin{document}
\maketitle
\begin{abstract}
An agent can spend almost as much deciding what to do as doing it. In
our documentation workflow, control-flow routing accounts for 42.5\% of
baseline reference cost on the initial task. Lower token prices alone do
not solve this control-flow tax: inexpensive general-purpose routers can
trigger additional calls to the expensive generator. We evaluate a
heterogeneous architecture in which Jev, a specialized System 1 decision
model, controls finite actions while Sol generates documents. Across
four software tasks, 480 valid runs comprise 470 completions and ten
request-limit endpoints. Jev full routing achieves the lowest mean cost
and elapsed time on every task, completing all 40 specialist-controlled
runs along the shortest trajectory. On the initial task, mean cost falls
by 49.6\% (95\% paired-bootstrap interval: 42.9\% to 56.3\%) and elapsed
time by 33.2\% (19.6\% to 46.7\%). Across three further tasks, cost
reductions are 40.5\%--42.9\% and concurrent elapsed-time reductions are
12.9\%--14.4\%. Existence-only specialization saves just 1.8\% on the
initial task. Separately, Sol and Jev each select accepted actions on
100/100 reviewed retail states, with Jev using 97.88\% less reference
cost and 72.30\% less mean request time. The results identify two
requirements for efficient agent control: inexpensive decisions and
trajectories that preserve those savings.
\end{abstract}

\section{Introduction}\label{introduction}

Agent workflows pay a recurring control-flow tax. Producing an artifact
is followed by another decision about whether to advance, repair,
backtrack, or stop. In a content-aware controller, selecting a short
action label can require rereading the full request, current files,
dependency contracts, and execution history. The answer is small; the
evidence needed to choose it is not. As artifacts accumulate, repeatedly
sending that evidence to a general-purpose model makes control an
expensive part of execution rather than negligible orchestration
overhead.

Our shortest three-document workflow requires seven control requests and
only three generation requests. In the matched Sol controls for D01 Jev
full routing, control accounts for 42.5\% of baseline reference cost and
66.5\% of input tokens. It also occupies 19.4\% of baseline elapsed
time, measured as the sum of routing-request durations divided by
session time. These are different burdens: routing consumes nearly half
the reference cost, not half the waiting time. Moreover, a routing
choice can add another expensive generation call, amplifying its
downstream cost beyond the selector request itself.

Two tempting shortcuts miss this problem. First, replacing only a small
Boolean gate leaves the expensive semantic chooser untouched: a 97.6\%
gate-cost reduction translates into just 1.8\% lower workflow cost on
D01. Second, choosing the cheapest general-purpose router can be a false
economy. With the same Sol generator, Haiku reaches the 80-request limit
in every D01 full-routing run, while Luna uses 54\% more requests than
its matched controls. Jev instead combines low decision overhead with
minimum-length completed trajectories. The central question is therefore
not whether a small model can emit a valid action, but whether it can
control the expensive work that follows.

\begin{figure}
\centering
\includegraphics[width=1\linewidth,height=\textheight,keepaspectratio,alt={Workflow-control
specialization. (a) A conceptual view of evidence, finite actions,
unchanged Sol generation, and artifact feedback; Tasks illustrates one
selected action, not a fixed schedule. Control combines next-step
selection and existence checking; Figure 2 separates these operations.
(b) Mean reference cost for D01 E2 Jev and its matched Sol controls,
decomposed into control and generation requests on a common scale.
Control accounts for 42.5\% of baseline cost; total reduction is 49.6\%
{[}42.9, 56.3{]} with a pointwise paired-bootstrap 95\% interval. The
same generator can incur different costs as decisions and trajectories
change.}]{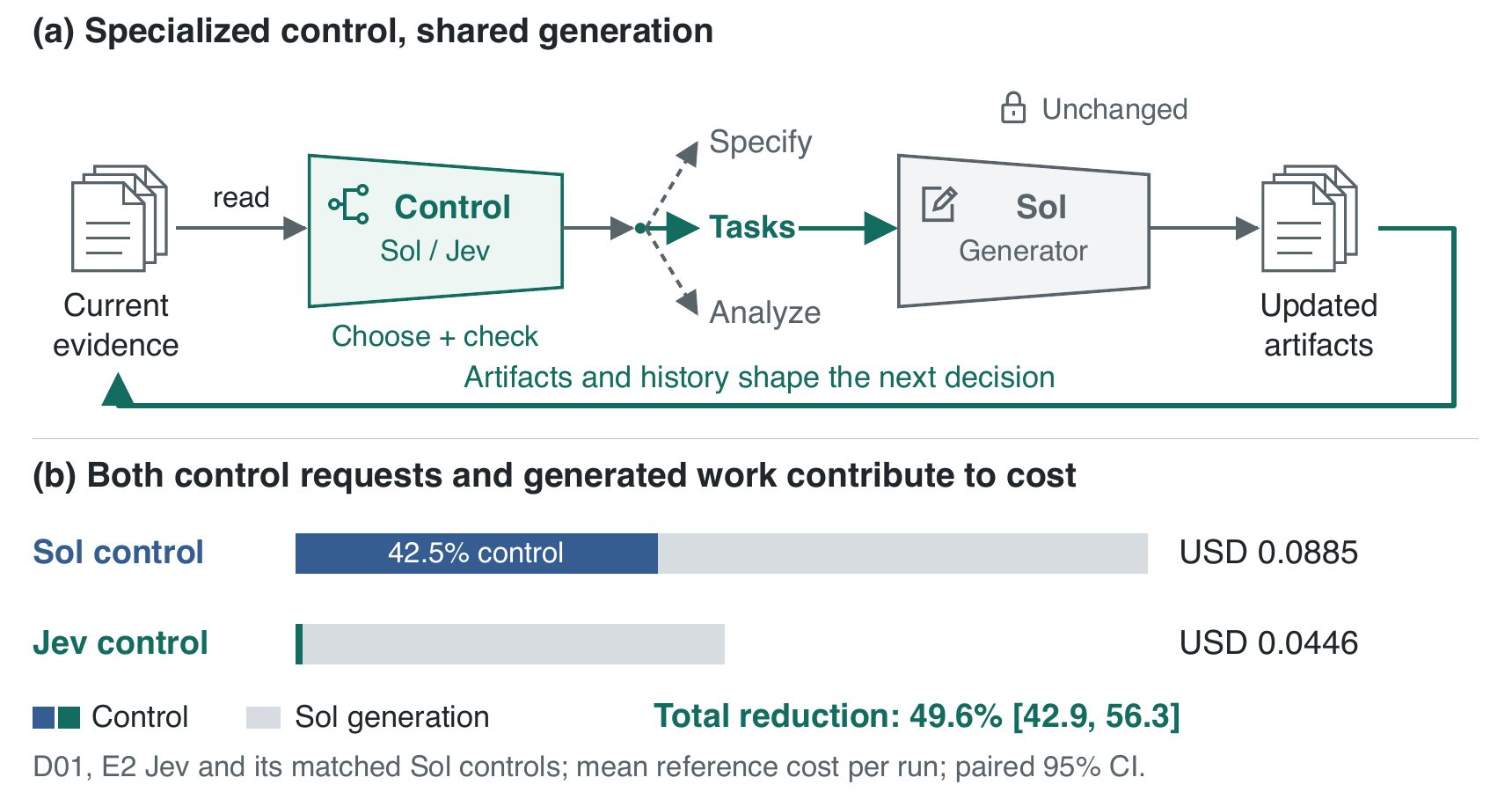}
\caption{Workflow-control specialization. (a) A conceptual view of
evidence, finite actions, unchanged Sol generation, and artifact
feedback; Tasks illustrates one selected action, not a fixed schedule.
Control combines next-step selection and existence checking; Figure 2
separates these operations. (b) Mean reference cost for D01 E2 Jev and
its matched Sol controls, decomposed into control and generation
requests on a common scale. Control accounts for 42.5\% of baseline
cost; total reduction is 49.6\% {[}42.9, 56.3{]} with a pointwise
paired-bootstrap 95\% interval. The same generator can incur different
costs as decisions and trajectories change.}
\end{figure}

Specialization motivates heterogeneous agent systems {\citep{ref1},}
while model routing {\citep{ref2},} cascades {\citep{ref3},} and
multi-agent routing {\citep{ref4}} optimize different deployment
choices. Our intervention keeps the document-generating model fixed and
changes the backend that selects the next workflow action. Routing here
means control-flow selection and its associated existence gate, not
selection of the final-answer model. Following application-level
evaluation concerns {\citep{ref5}} and cost-quality routing evaluation
{\citep{ref8},} we measure the entire resulting session, including the
downstream generation induced by its decisions.

We address three research questions: (RQ1) how does decision-backend
specialization affect observed routing cost and client-side request
duration; (RQ2) how do these differences translate into session cost,
elapsed time, and completion; and (RQ3) how does trajectory variation
affect the interpretation of local efficiency gains? We hold initial
conditions fixed and allow subsequent trajectories to diverge,
preserving the downstream consequences of each routing policy.

Our contributions are (1) a finite-choice control interface and shared
content-aware policy separating decision semantics from provider
encoding; (2) an existence-only versus full-routing comparison that
localizes the economically important specialization boundary; (3)
trajectory-inclusive comparisons against inexpensive general-purpose
routers across four software tasks; and (4) a same-state
customer-service evaluation that tests accepted action selection
independently of workflow evolution. Together, these studies support an
architecture in which specialized models handle frequent bounded
judgments and general-purpose models retain open-ended generation.

\section{Related Work}\label{related-work}

\textbf{Specialized decisions and intervention.} Belcak et
{al.~\citep{ref1}} argue that repeated, narrowly scoped agent
invocations can benefit from specialized models, while general-purpose
capabilities remain valuable elsewhere in the system. SOMA
{\citep{ref19}} uses a System 1 monitor to inspect an ongoing System 2
reasoning trace and issue reminders or take over after accumulated
evidence crosses a calibrated threshold. Its reported token accounting
covers the System 2 reasoner, and it does not measure wall-clock
latency; its efficiency results therefore are not directly comparable to
our full-session measurements. Decision4Rec {\citep{ref20}} combines
typed decision models with candidate generation, ranking, and score
fusion for recommendation. Its principal evaluation uses a batch
pipeline, and trained open decision models can outperform the released
Jev model. These studies motivate separating decision and generation
roles. Section 3.4 distinguishes Jev's published design from the
behavior measured here; System 1 and System 2 name computational roles,
not human cognition.

\textbf{Answer-model routing and cascades.} RouteLLM {\citep{ref2}}
learns to route requests between stronger and weaker generators using
preference data. FrugalGPT {\citep{ref3}} combines models in cost-aware
cascades. RouterBench {\citep{ref8}} evaluates predictive routing and
cascading using stored model responses, quality scores, and monetary
costs, including comparisons with random mixtures and oracle routing.
These approaches select the model used to answer a request; our
controller instead selects the next workflow action while retaining a
common generator. This distinction motivates measuring downstream
generation and repair in addition to selector overhead. RouterBench
defers latency and throughput evaluation, whereas we report
client-observed elapsed time under explicitly described serial or
concurrent execution. We do not implement these papers' learned routers,
oracle baselines, or a fallback from Jev to Sol.

\textbf{Reasoning, tools, and modular execution.} ReAct {\citep{ref9}}
interleaves reasoning, actions, and observations; Toolformer
{\citep{ref10}} learns when and how to invoke external APIs. Both place
tool use within a broader generative policy. ReWOO {\citep{ref12}}
separates planning, tool execution, and synthesis to reduce repeated
context, but its up-front plan is formed before tool observations. Our
chooser instead re-examines current artifact contents after execution.
LLMCompiler {\citep{ref11}} schedules dependencies between function
calls to exploit parallel execution within a task. Our expansion's
concurrency is different: independent experimental pairs share gateway
load, while each workflow remains sequential. These distinctions
separate controller specialization from prompt compression,
tool-argument generation, and task scheduling as potential sources of
efficiency.

\textbf{Feedback, search, and coordination.} Reflexion {\citep{ref13}}
uses verbal feedback and episodic memory to improve subsequent attempts,
while Tree of Thoughts {\citep{ref14}} explores intermediate reasoning
states using search and backtracking. Our repeat and backtrack actions
operate over three registered document-producing steps; they do not
implement learned reflective memory or thought-tree search. MasRouter
{\citep{ref4}} considers collaboration structures, agent roles, and
model selection. We instead fix the registered contracts and generation
backend, allowing only the action sequence to vary. Dynamic
orchestration is not inherently preferable: Agentless {\citep{ref18}}
uses a prescribed localization, repair, and validation pipeline for
software issues. Our E1/E2 comparison tests specialization boundaries
within one dynamic controller, not dynamic control against an optimized
fixed pipeline.

\textbf{Evaluation and software-task boundaries.} Kapoor et
{al.~\citep{ref5}} emphasize joint evaluation of cost and accuracy,
meaningful baselines, reproducibility, and appropriate holdout sets. We
measure full sessions and retain request-limit outcomes, but do not
independently establish artifact-quality equivalence. API-Bank
{\citep{ref15}} evaluates planning, API retrieval, and calls,
illustrating why selecting an action identity alone does not establish
correct tool use. SWE-bench {\citep{ref16}} evaluates repository-level
issue resolution, and SWE-agent {\citep{ref17}} studies interfaces for
navigation, editing, and testing. Our software tasks produce
documentation rather than executable repairs; controller completion must
not be read as a SWE-bench-style resolved-issue score. These evaluations
motivate stronger future outcome measures without turning unmeasured
baselines into experimental results.

\textbf{Customer service and statistical interpretation.} \(\tau\)-bench
{\citep{ref6}} evaluates tool-using agents interacting with simulated
users under domain policies, and \(\tau^2\)-bench {\citep{ref7}} extends
this setting with dual-control environments. We derive single retail
decision points with identical visible evidence and candidate actions
across backends, scoring accepted action identities rather than official
benchmark rollouts. Following the design-sensitive evaluation principles
discussed by Dror et {al.~\citep{ref21},} our workflow intervals
preserve paired arms and execution-order strata. Section 5.9 defines
their interpretation.

\section{System and Decision Policy}\label{system-and-decision-policy}

\subsection{A Documentation-Only Agent
Workflow}\label{a-documentation-only-agent-workflow}

We implement our controller using an isolated workflow runtime and a
provider-compatible model gateway. We evaluate one workflow with three
registered steps (Table 1). Each session produces a specification, a
task decomposition, and a consistency analysis for a requested software
change. We restrict execution to these documentation artifacts;
application implementation and business-test execution are not part of
the task.

\begin{table}[!htbp]
\centering
\caption{Registered steps and responsibilities.}\label{tab:1}
\small\setlength{\tabcolsep}{4pt}\renewcommand{\arraystretch}{1.12}
\sbox{\papertable}{\begin{tabular*}{\linewidth}{@{\extracolsep{\fill}}>{\raggedright\arraybackslash}p{\dimexpr 0.14\linewidth-2\tabcolsep\relax}>{\raggedright\arraybackslash}p{\dimexpr 0.28\linewidth-2\tabcolsep\relax}>{\raggedright\arraybackslash}p{\dimexpr 0.23\linewidth-2\tabcolsep\relax}>{\raggedright\arraybackslash}p{\dimexpr 0.35\linewidth-2\tabcolsep\relax}@{}}
\toprule
\textbf{Step} & \textbf{Required input} & \textbf{Required output} & \textbf{Responsibility} \\
\midrule
Specify & Original request and initial workspace & Specification & Define requirements and constraints \\
Tasks & Usable specification & Task document & Decompose implementation and planned verification work \\
Analyze & Usable specification and tasks & Analysis report & Check consistency, coverage, and unresolved issues \\
\bottomrule
\end{tabular*}}
\typeout{PAPER-TABLE-WIDTH:1:\the\wd\papertable:\the\linewidth}
\usebox{\papertable}
\end{table}

We limit each final document to 40 lines and isolate each session in a
fresh workspace. Completion refers to this documentation workflow, not
to delivery of the requested software. We preserve the same output
contracts and generation instructions across decision backends.

\subsection{Dynamic Execution}\label{dynamic-execution}

At each selection point, we provide the original request, task scope,
registered contracts, sorted directory inventory, complete current UTF-8
file contents, and execution history. We bind snapshots to file digests
and reject oversized, binary, symlinked, or out-of-scope evidence rather
than truncate it. The Python controller exposes the registered steps,
\texttt{complete}, and explicit \texttt{abstain}.

We execute the selected step with Sol, then evaluate an existence gate
using only required output paths and Boolean existence values. A
positive gate returns control to the content-aware chooser; a negative
gate requests another execution of the same step. The chooser may repeat
a completed step or return to an earlier prerequisite. Every selection,
generation, and gate request consumes the same run-local allowance of 80
requests; the Python study adds no separate round or repair-count limit.
Figure 2 summarizes the execution loop and specialization boundaries.

\begin{figure}
\centering
\includegraphics[width=1\linewidth,height=\textheight,keepaspectratio,alt={Execution
loop and specialization boundaries. (a) A content-aware chooser reads a
scope- and digest-checked snapshot, dispatches a registered step to Sol,
or requests complete or abstain. The gate receives required paths and
Boolean existence values: true refreshes the evidence; false repeats
generation of the same step. Completion additionally requires the
deterministic terminal constraints shown. Invalid decisions, abstention,
and request exhaustion stop execution. Icons distinguish evidence, model
calls, artifacts, and deterministic safeguards; an existence check does
not certify artifact quality. (b) E1 changes only the gate backend; E2
changes both decision backends. Sol generation and the independent
80-request allowance per arm remain unchanged.}]{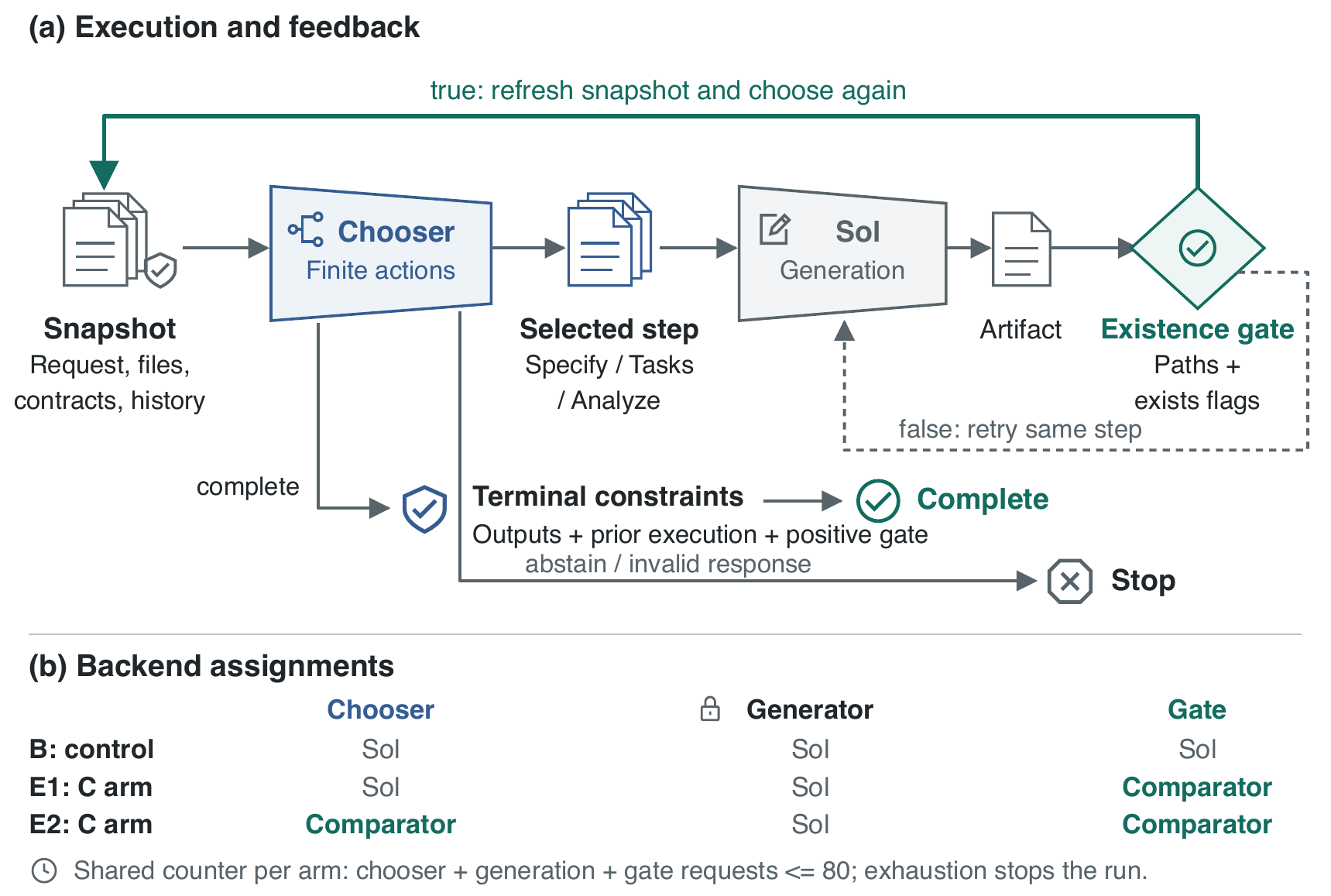}
\caption{Execution loop and specialization boundaries. (a) A
content-aware chooser reads a scope- and digest-checked snapshot,
dispatches a registered step to Sol, or requests complete or abstain.
The gate receives required paths and Boolean existence values: true
refreshes the evidence; false repeats generation of the same step.
Completion additionally requires the deterministic terminal constraints
shown. Invalid decisions, abstention, and request exhaustion stop
execution. Icons distinguish evidence, model calls, artifacts, and
deterministic safeguards; an existence check does not certify artifact
quality. (b) E1 changes only the gate backend; E2 changes both decision
backends. Sol generation and the independent 80-request allowance per
arm remain unchanged.}
\end{figure}

We reject completion when required outputs are absent, no step has
executed, or the latest execution lacks a positive existence gate. These
deterministic constraints complement, but do not independently verify,
the chooser's content assessment.

\subsection{Shared Content-Aware
Prompt}\label{shared-content-aware-prompt}

We use the same content-aware policy and candidate meanings across
Python comparison conditions. Our policy separates trusted scope and
contracts from generated evidence and defines the chooser as a decision
maker rather than a writer or tool executor. It checks actual artifact
contents, repairs the earliest unresolved prerequisite, and permits
completion only after all requirements are satisfied. Appendix A.1 gives
the five decision rules.

We map Jev's Choice outputs \texttt{choice\_0} through
\texttt{choice\_3} to Specify, Tasks, Analyze, and complete, with
\texttt{abstain} as a separate rejection outcome. General-purpose
backends use a schema-constrained choice over the same candidates. Sol
and Luna use the Responses API; Haiku uses an explicitly configured Chat
Completions API. We validate model identity, response structure, and
usage; for Jev, we additionally check probability validity and agreement
with the selected maximum. We allow bounded probability-rounding error
without modifying the returned distribution. Invalid responses and
abstention terminate the run. There is no automatic provider
substitution.

Our comparison equalizes decision policy, evidence content, candidate
meaning, and evidence limits, while retaining provider-specific
transport formats. We verify equality of the initial decision inputs.
Later inputs may differ because each condition independently generates
documents and chooses subsequent actions.

\subsection{Why a Decision Model Is
Different}\label{why-a-decision-model-is-different}

A decision model is defined by its role, not necessarily by a
non-autoregressive architecture: a language model can also make
decisions by generating constrained action tokens. Jev takes a different
output path. TypeSafe's technical introduction describes a dedicated
architecture, a parallel sampler that emits probabilities together
rather than generating text token by token, and Reinforcement Learning
for Calibrated Decisions (RLCD) {\citep{ref22}.} Thus Jev's published
decision-output mechanism is non-autoregressive; this does not establish
an encoder-only backbone or a specific parameter count, neither of which
is disclosed in that account.

The distinction from a cheaper chat model is substantive. Jev accepts
natural-language evidence but gives up open-ended string generation in
favor of typed choices and distributions over declared alternatives.
JSON-constrained language models restrict the format of generated text;
the published Jev design instead specializes the output mechanism and
training objective for decisions. This creates a plausible route to
low-cost control without discarding the evidence needed for a difficult
choice. Our general-purpose baselines already use structured choices, so
the comparison is not merely typed output versus unconstrained prose.

The practical advantage still has to survive execution. A low-cost
selector that repeatedly reactivates Sol can lose its savings, and a
schema-valid choice can be semantically wrong. We therefore evaluate
both the induced workflow trajectory and, separately, accepted actions
on matched retail states. Jev's returned probabilities are validated
structurally; their empirical calibration is not measured in this study.

\section{Experimental Design and
Measurement}\label{experimental-design-and-measurement}

\subsection{Common-Runtime
Comparisons}\label{common-runtime-comparisons}

\begin{table}[!htbp]
\centering
\caption{Actual backend assignments. Generation is Sol in every cell.}\label{tab:2}
\small\setlength{\tabcolsep}{4pt}\renewcommand{\arraystretch}{1.12}
\sbox{\papertable}{\begin{tabular*}{\linewidth}{@{\extracolsep{\fill}}>{\raggedright\arraybackslash}p{\dimexpr 0.28\linewidth-2\tabcolsep\relax}>{\raggedright\arraybackslash}p{\dimexpr 0.08\linewidth-2\tabcolsep\relax}>{\raggedright\arraybackslash}p{\dimexpr 0.29\linewidth-2\tabcolsep\relax}>{\raggedright\arraybackslash}p{\dimexpr 0.35\linewidth-2\tabcolsep\relax}@{}}
\toprule
\textbf{Experiment} & \textbf{Arm} & \textbf{choose\_step} & \textbf{Existence gate} \\
\midrule
E1: existence-only & B & Sol & Sol \\
E1: existence-only & C & Sol & Selected comparator \\
\midrule
E2: full routing & B & Sol & Sol \\
E2: full routing & C & Selected comparator & Selected comparator \\
\bottomrule
\end{tabular*}}
\typeout{PAPER-TABLE-WIDTH:2:\the\wd\papertable:\the\linewidth}
\usebox{\papertable}
\end{table}

The models are Sol (\texttt{gpt-6.1-sol}), Jev (\texttt{jev-1.13.0}),
Haiku (\texttt{claude-haiku-5.5}), Luna (\texttt{gpt-6-luna}), and
Sonnet (\texttt{claude-sonnet-5.5}). Sol supplies every generator and
all B decisions. D01 compares Jev, Haiku, and Luna; Sonnet appears only
in the expansion. We collect ten matched B/C pairs for each comparator
and specialization boundary on D01, giving six conditions, 60 pairs, and
120 valid runs. Every condition has fresh B controls. A run is one
workflow arm, not one API request. Initial inputs, generation policy,
controller semantics, and limits are shared apart from the decision
backend and its transport encoding. Different trajectories are part of
each system's measured behavior. Paired reductions use each comparator's
matched Sol controls; absolute means also compare C backends directly.

\subsection{Task, Repetitions, and
Ordering}\label{task-repetitions-and-ordering}

We use one synthetic development task, D01, concerning a
queue-normalization library. Its requirements cover identifier and count
validation, duplicate rejection, nonmutating sorting, field selection,
and preservation of a separate UI service. We evaluate documentation of
these requirements rather than executable implementation. We inspected
this task during system development; it is not a held-out benchmark.

We instantiate the same request and initial files separately for every
run and do not share generated documents or histories between arms. Each
condition has five B-then-C and five C-then-B pairs, executed
sequentially according to a frozen schedule. We do not control
generation seeds. The balanced order distributes first- and second-run
positions equally between backends.

The analysis contains one valid observation for each planned arm and
repetition. A valid experimental execution reaches either controller
completion or the 80-request endpoint, with resource measurements
available. Both terminal outcomes remain in the analysis. Collection
provenance and cohort construction are documented in the accompanying
reproducibility note.

\subsection{Runtime and Measurement}\label{runtime-and-measurement}

The primary workflow executor uses Python 3.11.12, HTTPX 0.28.1, and
Pydantic 2.14.1. Provider receipts verify the configured model
identifiers. Each arm may transmit at most 80 requests across all
stages; request 81 is rejected before transmission. The B and C counters
are independent. We set no additional experimental time, aggregate-cost,
output-token, round, or repair-count budget. Provider limits and strict
document, schema, filesystem, and request/response-byte safeguards still
apply. This configuration is not an unlimited capability test.

We report resource use for valid experimental runs, including all
requests made before their terminal outcome. Costs below therefore
measure the evaluated workflow executions, rather than collection-level
expenditure.

For receipt-verified request usage, reference cost is

\[\hat{C}_r = \frac{p_{\mathrm{in}}t_{r,\mathrm{in}} + p_{\mathrm{out}}t_{r,\mathrm{out}}}{10^6}.\]

We use fixed reference tariffs per million input/output tokens of
USD2/10 for Sol and Sonnet, USD0.042/0 for Jev, and USD0.10/0.50 for
both Haiku and Luna. These tariffs define a reproducible cost metric
rather than measured invoices. We count every metered generation,
selection, existence check, and repair once, without applying cache
discounts.

Let \(G\), \(Q\), and \(E\) be disjoint sets of generation, next-step,
and existence requests. The model cost and routing subtotal are

\[\begin{aligned}
\hat{C}_{\mathrm{routing}} &= \sum_{r\in Q\cup E}\hat{C}_r,\\
\hat{C}_{\mathrm{session}} &= \sum_{r\in G}\hat{C}_r + \hat{C}_{\mathrm{routing}}.
\end{aligned}\]

Our generation category includes all metered document-producing turns.
We report per-request and per-session model costs; we do not estimate
local computation, labor, model training, or deployment amortization.

We distinguish elapsed session time from
\(T_{\mathrm{routing}}=\sum_{r\in Q\cup E}d_r\), the sum of client-side
request durations. The latter includes transport and gateway overhead
and is neither pure model inference time nor time to first token. Means
of per-session sums and per-request means have different denominators
when request counts differ. Missing durations remain missing, not zero.

\subsection{Outcome Measures}\label{outcome-measures}

We measure valid experimental execution, controller completion, selected
step sequences, request counts, per-stage reference costs, elapsed time,
and artifact integrity. Reaching the request limit is a successful
observation of an expected experimental endpoint and remains in every
resource summary. We separately label whether the workflow completed or
reached the cap: experimental validity and workflow completion are
distinct outcomes. Elapsed time measures time to the observed endpoint.
We report arithmetic means and pointwise paired-bootstrap intervals,
with the resampling unit specified in Section 5.9 and Appendix C.

The specialized-decision subtotal means existence checks alone in E1 and
both next-step and existence requests in E2. This boundary-specific
measure keeps unchanged Sol chooser costs in E1 separate from comparator
costs. The workflow outcome measures emphasize execution and resource
use; retail evaluation additionally scores reviewed action selection.

\subsection{Single-Decision Customer-Service
Evaluation}\label{single-decision-customer-service-evaluation}

The reviewed retail study isolates one routing decision at a time. We
author 100 counterfactual decision states from 30 source tasks and 25
customer accounts in the test-account pool of \(\tau^2\)-bench
{\citep{ref7},} release v1.0.1. Each account contributes four states.
These are not 100 natural conversations, independent source tasks, or
official benchmark labels. The visible input contains customer turns,
prior tool evidence, the retail policy, and 17 candidate actions: reply
or one of 16 tools. Labels, source identifiers, and scoring criteria are
excluded from requests. Source-account separation from the development
pilot does not imply scenario-family independence.

The acceptable-action sets contain 45 reply-only cases, 52 tool-only
cases, and three arithmetic cases accepting either calculate or reply.
All 16 tools occur as primary actions in at least three cases.
Construction includes 55 isolated native action checks; these validate
fixture behavior, not independent semantic labels. The project lead
reviews the complete set before live execution, with a signed review
record and retained raw annotations. This is single-reviewer
project-side assessment, not independent blinded adjudication.

Sol and Jev each make one request per case on identical visible evidence
and candidate meanings, following a frozen order with no retries or
fallback. The unit scored is membership in the pre-run acceptable-action
set, not exact agreement between backends, tool-argument correctness, or
completed customer resolution. The older 20-case development pilot
remains separate, as documented in the accompanying reproducibility
note.

\subsection{Evaluation Dataset and Sampling
Protocol}\label{evaluation-dataset-and-sampling-protocol}

We evaluate four software requirements tasks and one retail decision
dataset. D01 concerns queue normalization. Three further fixtures cover
D02, JSON Merge Patch; D03, a bounded Retry-After application profile;
and D04, Semantic Version precedence. Together they test data
transformation, HTTP timing policy, and version-ordering requirements
within the same documentation workflow. Retail adds a distinct
customer-service domain, with policy-constrained choices among reply and
16 tools.

The D02--D04 expansion retains Sol generation and controls, Jev and
Luna, and uses Sonnet as its third comparator. It contains ten pairs per
task and condition: 180 pairs and 360 valid runs, all ending in workflow
completion. Sonnet was chosen after the initial Haiku evaluation; the
D01 Haiku observations remain a separate baseline. Sonnet uses an
auto-tool Chat interface, with a strict adapter validating the
structured decision independently of explanatory text. All models share
the same candidate meanings and content-aware policy.

The expansion executes interleaved batches of up to 12 isolated pairs,
retaining sequential BC/CB order inside each pair. Timings measure
concurrent execution under the observed gateway load. We report each
task-condition cell separately and keep these timings distinct from
serial D01 measurements.

\section{Results}\label{results}

\subsection{Existence-Only Savings Hit a Workflow
Ceiling}\label{existence-only-savings-hit-a-workflow-ceiling}

All 60 selected E1 runs complete. Jev reduces existence-check cost by
97.6\%, yet full-run cost falls by only 1.8\% (Table 3). The gate
accounts for just 3.0\% of matched baseline cost. This is an
Amdahl-style constraint: holding the remaining work fixed, the measured
gate saving would remove only about 2.9\% of baseline session cost.
Small changes in the much larger generation and chooser components can
readily outweigh that gain. Haiku's full-cost reduction is 4.1\%; Luna's
point estimate favors the Sol control.

\begin{table}[!htbp]
\centering
\caption{E1 reductions relative to each comparator’s matched Sol controls (ten pairs per row). Intervals are pointwise 95\% paired-bootstrap CIs; negative reductions mean increases. Higher reductions are better. Bold marks the largest observed reduction per metric.}\label{tab:3}
\small\setlength{\tabcolsep}{4pt}\renewcommand{\arraystretch}{1.12}
\sbox{\papertable}{\begin{tabular*}{\linewidth}{@{\extracolsep{\fill}}lrrrrrr@{}}
\toprule
 & \multicolumn{2}{c}{Full cost reduction (\%) $\uparrow$} & \multicolumn{2}{c}{Elapsed reduction (\%) $\uparrow$} & \multicolumn{2}{c}{Gate cost reduction (\%) $\uparrow$} \\
\cmidrule(lr){2-3}\cmidrule(lr){4-5}\cmidrule(lr){6-7}
Backend & Est. & 95\% CI & Est. & 95\% CI & Est. & 95\% CI \\
\midrule
Haiku & {\color{PaperTeal}\bfseries 4.1} & \textcolor{PaperMuted}{$[0.8,\,7.1]$} & {\color{PaperTeal}\bfseries 7.3} & \textcolor{PaperMuted}{$[4.1,\,10.0]$} & 87.6 & \textcolor{PaperMuted}{$[87.6,\,87.6]$} \\
Jev & 1.8 & \textcolor{PaperMuted}{$[-0.9,\,4.4]$} & 3.9 & \textcolor{PaperMuted}{$[1.1,\,6.6]$} & {\color{PaperTeal}\bfseries 97.6} & \textcolor{PaperMuted}{$[97.6,\,97.6]$} \\
Luna & -3.8 & \textcolor{PaperMuted}{$[-20.3,\,5.8]$} & -1.4 & \textcolor{PaperMuted}{$[-12.2,\,6.6]$} & 93.4 & \textcolor{PaperMuted}{$[92.7,\,93.9]$} \\
\bottomrule
\end{tabular*}}
\typeout{PAPER-TABLE-WIDTH:3:\the\wd\papertable:\the\linewidth}
\usebox{\papertable}
\end{table}

The narrow gate-cost intervals reflect repeated small contracts and
displayed rounding. The main optimization target is elsewhere: the
content-aware chooser still rereads accumulated evidence through Sol. E2
therefore specializes both control boundaries.

\subsection{Full Routing Exposes the Cheap-Generalist
Trap}\label{full-routing-exposes-the-cheap-generalist-trap}

Specializing both next-step selection and existence checking produces
markedly different trajectories across backends. Jev reduces mean
full-run reference cost by 49.6\% and elapsed time by 33.2\%, with both
arms completing all ten workflows. Luna also completes all workflows,
but its full-run intervals include substantial benefit and harm. Haiku
reaches the 80-request cap in all ten C runs. These ten valid
experimental outcomes remain in the resource comparison and are labeled
as request-limit endpoints rather than workflow completions.

\begin{table}[!htbp]
\centering
\caption{E2 reductions relative to matched Sol controls (ten pairs per row). Intervals are pointwise 95\% paired-bootstrap CIs. Routing includes selection and existence checking. Higher reductions are better; bold marks the largest observed reduction per metric. Haiku includes ten request-limit endpoints.}\label{tab:4}
\small\setlength{\tabcolsep}{4pt}\renewcommand{\arraystretch}{1.12}
\sbox{\papertable}{\begin{tabular*}{\linewidth}{@{\extracolsep{\fill}}lrrrrrr@{}}
\toprule
 & \multicolumn{2}{c}{Full cost reduction (\%) $\uparrow$} & \multicolumn{2}{c}{Elapsed reduction (\%) $\uparrow$} & \multicolumn{2}{c}{Routing cost reduction (\%) $\uparrow$} \\
\cmidrule(lr){2-3}\cmidrule(lr){4-5}\cmidrule(lr){6-7}
Backend & Est. & 95\% CI & Est. & 95\% CI & Est. & 95\% CI \\
\midrule
Haiku & -734.3 & \textcolor{PaperMuted}{$[-759.5,\,-709.5]$} & -591.7 & \textcolor{PaperMuted}{$[-635.5,\,-542.8]$} & 2.8 & \textcolor{PaperMuted}{$[-0.4,\,5.5]$} \\
Jev & {\color{PaperTeal}\bfseries 49.6} & \textcolor{PaperMuted}{$[42.9,\,56.3]$} & {\color{PaperTeal}\bfseries 33.2} & \textcolor{PaperMuted}{$[19.6,\,46.7]$} & {\color{PaperTeal}\bfseries 98.1} & \textcolor{PaperMuted}{$[97.8,\,98.3]$} \\
Luna & -3.7 & \textcolor{PaperMuted}{$[-75.3,\,37.7]$} & -53.6 & \textcolor{PaperMuted}{$[-157.4,\,7.7]$} & 87.9 & \textcolor{PaperMuted}{$[78.9,\,93.3]$} \\
\bottomrule
\end{tabular*}}
\typeout{PAPER-TABLE-WIDTH:4:\the\wd\papertable:\the\linewidth}
\usebox{\papertable}
\end{table}

Haiku and Luna have the same low reference tariff, yet neither saves
full-run cost on D01 at the point estimate. Haiku's repeated decisions
almost erase even its routing-cost advantage, while additional Sol
generation raises total cost by 734.3\%. Luna retains an 87.9\%
routing-cost reduction but incurs a 3.7\% increase in full-run cost. Jev
avoids this false economy: routing cost falls by 98.1\%, and the saving
survives at the session level. The differentiator is the combination of
decision price and execution behavior, not the low tariff alone.

\begin{figure}
\centering
\includegraphics[width=1\linewidth,height=\textheight,keepaspectratio,alt={Full-run
reference-cost reductions on D01, with pointwise 95\% paired-bootstrap
intervals. Negative values denote increases. Haiku E2 includes ten valid
request-limit outcomes, with a separate scale.}]{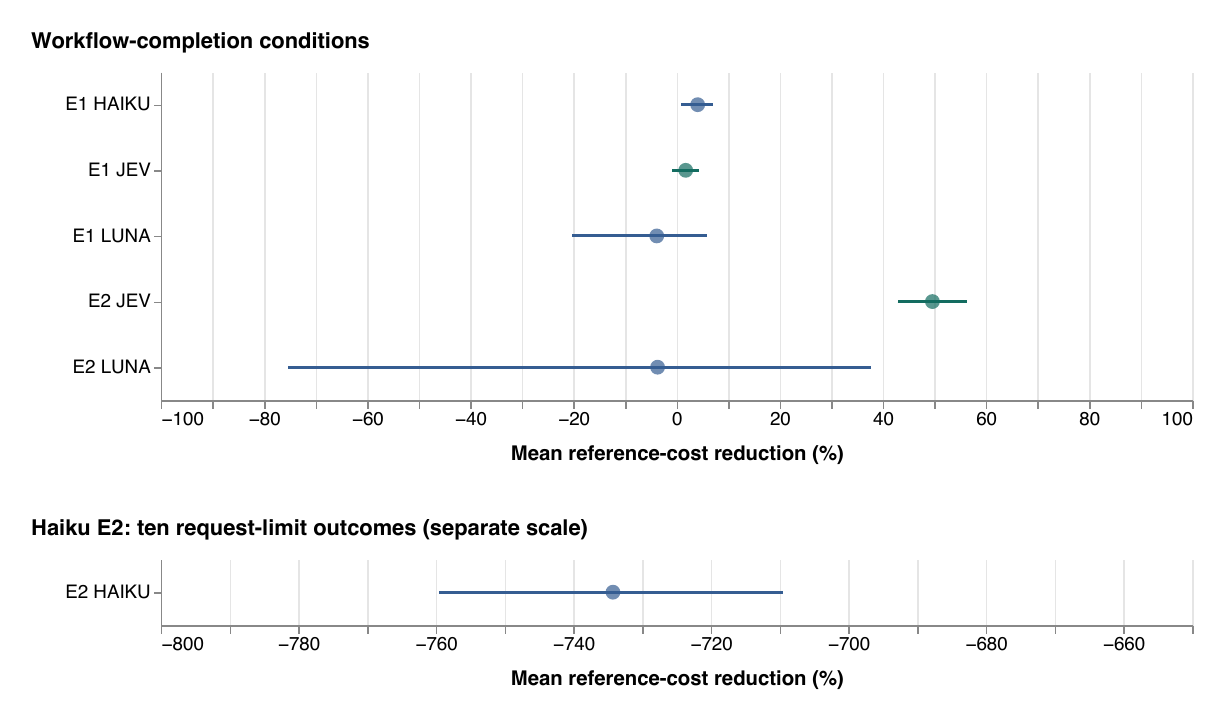}
\caption{Full-run reference-cost reductions on D01, with pointwise 95\%
paired-bootstrap intervals. Negative values denote increases. Haiku E2
includes ten valid request-limit outcomes, with a separate scale.}
\end{figure}

Appendix B.1 reports all D01 arm-level resources (Table 8). Jev E2 has
the lowest observed mean cost and elapsed time across the D01
configurations. Fresh controls support within-condition paired
reductions, while absolute means summarize the deployed systems'
end-to-end behavior.

\subsection{Trajectories Determine the Price of
Control}\label{trajectories-determine-the-price-of-control}

The shortest completed trajectory is Specify, Tasks, Analyze, complete:
three generation requests, four chooser requests, and three existence
checks. Jev follows this trajectory in all 40 E2 runs across D01--D04,
with 10.0 requests per run and no repeated document generation. On D01,
its ten runs use 100 requests versus 109 for the matched Sol controls,
whose trajectories include three additional generation calls.

Luna's D01 total of 154 requests comprises 48 generations, 58 choices,
and 48 existence checks, compared with 30, 40, and 30 in its controls.
Thus 54\% more requests means 60\% more generation calls, not 54\% more
workflow steps. Haiku's ten capped runs make 270 generation calls versus
30 in their controls, repeatedly revisiting Specify or Tasks. Each cap
is 80 model requests, not 80 executed steps.

These traces explain why selector-only benchmarks can mislead: the
router also determines how often the expensive generator runs. They do
not require labeling every repeat as an error. Luna takes minimum-length
paths on D02 and D03 and achieves cost savings there, so the
cheap-generalist trap is a workload-dependent risk, not a claim that
inexpensive language models always fail. Jev's distinction in this study
is that low routing overhead and minimum-length completion occur
together on every evaluated task.

\subsection{Valid Experimental
Outcomes}\label{valid-experimental-outcomes}

All 120 D01 runs produce a valid terminal observation: 110 workflows
complete and ten reach the expected request limit. The three-task
expansion contributes another 360 valid runs, all with workflow
completion. Thus the workflow analysis contains 480 valid experimental
outcomes, including every request-capped run. Table 5 summarizes
resource use for these analyzed runs.

\begin{table}[!htbp]
\centering
\caption{Valid workflow outcomes and analyzed-run resource totals. Request-limit endpoints are valid experiments, distinct from workflow completion.}\label{tab:5}
\small\setlength{\tabcolsep}{4pt}\renewcommand{\arraystretch}{1.12}
\sbox{\papertable}{\begin{tabular*}{\linewidth}{@{\extracolsep{\fill}}lrr@{}}
\toprule
\textbf{Quantity} & \textbf{D01} & \textbf{D02–D04} \\
\midrule
Valid experimental runs & 120 & 360 \\
Workflow completions & 110 & 360 \\
Request-limit outcomes & 10 & 0 \\
Model requests & 1,969 & 3,732 \\
Reference cost, USD & 15.042193 & 37.667887 \\
\bottomrule
\end{tabular*}}
\typeout{PAPER-TABLE-WIDTH:5:\the\wd\papertable:\the\linewidth}
\usebox{\papertable}
\end{table}

\subsection{Single-Decision Agreement on Customer-Service
Cases}\label{single-decision-agreement-on-customer-service-cases}

Sol and Jev select a member of the pre-run reviewed acceptable-action
set on all 100 cases. Their choices agree exactly on 97 cases; the
remaining three choose different accepted arithmetic alternatives. All
200 requests are valid and fully accounted. Because both models see the
same policy, evidence, and candidates, this comparison isolates action
selection from the evolving artifacts and differing call counts of the
software workflows (Table 6).

\begin{table}[!htbp]
\centering
\caption{Reviewed retail decision study. Token and cost totals cover 100 requests per backend; duration is the arithmetic mean per request. Bold marks the lowest observed cost and duration; accepted-action counts are tied.}\label{tab:6}
\small\setlength{\tabcolsep}{4pt}\renewcommand{\arraystretch}{1.12}
\sbox{\papertable}{\begin{tabular*}{\linewidth}{@{\extracolsep{\fill}}lrr@{}}
\toprule
\textbf{Metric} & \textbf{Sol} & \textbf{Jev} \\
\midrule
Accepted actions & 100/100 & 100/100 \\
Input tokens & 381,316 & 397,647 \\
Output tokens & 2,505 & 18,821 \\
Total reference cost, USD $\downarrow$ & 0.787682 & {\color{PaperTeal}\bfseries 0.016701} \\
Mean request duration, seconds $\downarrow$ & 2.519 & {\color{PaperTeal}\bfseries 0.698} \\
\bottomrule
\end{tabular*}}
\typeout{PAPER-TABLE-WIDTH:6:\the\wd\papertable:\the\linewidth}
\usebox{\papertable}
\end{table}

Jev uses 97.88\% less reference cost and 72.30\% less mean request time.
Total reference cost is USD0.804383 and collection wall time is 322.105
seconds. Costs in Tables 5 and 6 use six decimal places, durations use
three, and percentage reductions are computed from unrounded
measurements. Matching all reviewed acceptable-action sets at this lower
cost extends the evidence beyond software-step selection to
policy-constrained customer service. The three accepted disagreements
also illustrate why exact backend agreement and acceptable-action
membership are different outcomes.

\subsection{Routing Outcomes and
Validation}\label{routing-outcomes-and-validation}

Workflow validation checks the required paths, ordered headings, line
limits, initial-file preservation, and terminal controller conditions.
Request-limit outcomes are separately identifiable, allowing trajectory
and resource comparisons without discarding expected endpoints. The
retail study adds a semantic decision measure: membership in
acceptable-action sets reviewed before execution.

The workflow checks establish execution integrity, while retail
contributes a reviewed semantic action measure. Together they test
complementary properties: whether the controller follows its execution
contract and whether a finite choice is acceptable on a given state. The
remaining outcome-quality boundary is discussed in Section 7.3.

\subsection{Cross-Task Generalization}\label{cross-task-generalization}

The expansion contains ten matched pairs in each of 18 task-condition
cells, for 360 valid workflow completions. Relative to its matched Sol
controls, Jev E2 reduces full-run cost on D02 by 41.6\% {[}40.4,
42.8{]}, D03 by 42.9\% {[}38.6, 47.8{]}, and D04 by 40.5\% {[}37.2,
45.0{]}. The 40.5\%--42.9\% reduction therefore recurs across JSON
transformation, HTTP timing policy, and version-ordering requirements.
Corresponding concurrent elapsed-time reductions are 12.9\%, 13.4\%, and
14.4\%. Table 7 reports absolute E2 resources alongside each backend's
cost reduction against its own matched Sol controls. Figure 4 displays
those controls and labels Jev's cost and time reductions directly.
Appendix B.2 retains all 18 paired contrasts and timing intervals (Table
9).

Jev E2 shows lower mean cost and elapsed time on each fixture, whereas
Luna E2 shows cost reductions without a corresponding latency reduction.
Sonnet E2 costs more at the point estimate on all three tasks, with D03
and D04 cost intervals excluding zero reduction. Its D02 elapsed-time
reduction coexists with higher mean cost. These direct measurements
distinguish cost efficiency from speed and include each backend's
resulting trajectory.

\par\medskip\noindent\begin{minipage}{\linewidth}
\centering
\captionof{table}{E2 absolute resources and cost reductions against each backend’s own matched Sol controls (ten pairs per row; pointwise 95\% CIs). Lower absolute means and higher reductions are better; negative reductions mean cost increases. Bold marks the lowest absolute mean within each task. D01 is serial; D02–D04 use concurrent load. Haiku D01 reaches the request limit; all other rows complete.}\label{tab:7}
\small\setlength{\tabcolsep}{3pt}\renewcommand{\arraystretch}{1.12}
\sbox{\papertable}{\begin{tabular*}{\linewidth}{@{\extracolsep{\fill}}lrrrrrr@{}}
\toprule
 & \multicolumn{2}{c}{Reference cost (USD) $\downarrow$} & \multicolumn{2}{c}{Elapsed time (s) $\downarrow$} & \multicolumn{2}{c}{Cost reduction vs Sol (\%) $\uparrow$} \\
\cmidrule(lr){2-3}\cmidrule(lr){4-5}\cmidrule(lr){6-7}
Backend & Mean & 95\% CI & Mean & 95\% CI & Est. & 95\% CI \\
\midrule
\addlinespace[3pt]\multicolumn{7}{l}{\color{PaperBlue}\bfseries D01} \\
Haiku & 0.6574 & \textcolor{PaperMuted}{$[0.6358,\,0.6797]$} & 596.2 & \textcolor{PaperMuted}{$[542.9,\,645.2]$} & -734.3 & \textcolor{PaperMuted}{$[-759.5,\,-709.5]$} \\
Jev & {\color{PaperTeal}\bfseries 0.0446} & \textcolor{PaperMuted}{$[0.0434,\,0.0459]$} & {\color{PaperTeal}\bfseries 73.3} & \textcolor{PaperMuted}{$[70.6,\,76.0]$} & 49.6 & \textcolor{PaperMuted}{$[42.9,\,56.3]$} \\
Luna & 0.0820 & \textcolor{PaperMuted}{$[0.0496,\,0.1379]$} & 146.4 & \textcolor{PaperMuted}{$[97.4,\,224.5]$} & -3.7 & \textcolor{PaperMuted}{$[-75.3,\,37.7]$} \\
\addlinespace[3pt]\multicolumn{7}{l}{\color{PaperBlue}\bfseries D02} \\
Jev & {\color{PaperTeal}\bfseries 0.0553} & \textcolor{PaperMuted}{$[0.0544,\,0.0562]$} & {\color{PaperTeal}\bfseries 83.4} & \textcolor{PaperMuted}{$[79.0,\,87.7]$} & 41.6 & \textcolor{PaperMuted}{$[40.4,\,42.8]$} \\
Luna & 0.0577 & \textcolor{PaperMuted}{$[0.0572,\,0.0583]$} & 102.9 & \textcolor{PaperMuted}{$[98.6,\,107.1]$} & 38.6 & \textcolor{PaperMuted}{$[37.7,\,39.4]$} \\
Sonnet & 0.1231 & \textcolor{PaperMuted}{$[0.1215,\,0.1248]$} & 91.9 & \textcolor{PaperMuted}{$[88.3,\,96.0]$} & -11.2 & \textcolor{PaperMuted}{$[-31.1,\,3.8]$} \\
\addlinespace[3pt]\multicolumn{7}{l}{\color{PaperBlue}\bfseries D03} \\
Jev & {\color{PaperTeal}\bfseries 0.0652} & \textcolor{PaperMuted}{$[0.0641,\,0.0662]$} & {\color{PaperTeal}\bfseries 97.1} & \textcolor{PaperMuted}{$[94.1,\,100.3]$} & 42.9 & \textcolor{PaperMuted}{$[38.6,\,47.8]$} \\
Luna & 0.0678 & \textcolor{PaperMuted}{$[0.0662,\,0.0694]$} & 123.1 & \textcolor{PaperMuted}{$[115.9,\,130.7]$} & 37.9 & \textcolor{PaperMuted}{$[36.4,\,39.6]$} \\
Sonnet & 0.1554 & \textcolor{PaperMuted}{$[0.1364,\,0.1907]$} & 113.9 & \textcolor{PaperMuted}{$[99.5,\,137.8]$} & -42.2 & \textcolor{PaperMuted}{$[-76.4,\,-24.0]$} \\
\addlinespace[3pt]\multicolumn{7}{l}{\color{PaperBlue}\bfseries D04} \\
Jev & {\color{PaperTeal}\bfseries 0.0666} & \textcolor{PaperMuted}{$[0.0655,\,0.0677]$} & {\color{PaperTeal}\bfseries 92.5} & \textcolor{PaperMuted}{$[87.3,\,97.7]$} & 40.5 & \textcolor{PaperMuted}{$[37.2,\,45.0]$} \\
Luna & 0.0851 & \textcolor{PaperMuted}{$[0.0741,\,0.0964]$} & 140.0 & \textcolor{PaperMuted}{$[122.5,\,159.4]$} & 30.4 & \textcolor{PaperMuted}{$[12.5,\,45.6]$} \\
Sonnet & 0.1545 & \textcolor{PaperMuted}{$[0.1357,\,0.1786]$} & 119.7 & \textcolor{PaperMuted}{$[101.8,\,139.5]$} & -39.7 & \textcolor{PaperMuted}{$[-60.7,\,-22.8]$} \\
\bottomrule
\end{tabular*}}
\typeout{PAPER-TABLE-WIDTH:7:\the\wd\papertable:\the\linewidth}
\usebox{\papertable}
\end{minipage}\par\medskip

\begin{figure}
\centering
\includegraphics[width=1\linewidth,height=\textheight,keepaspectratio,alt={E2
comparisons accompanying Table 7. Filled circles show each decision
backend; hollow diamonds show its own matched Sol controls. Eight
subplots retain absolute mean cost (left) and elapsed time (right), with
pointwise 95\% intervals. Subtitles report Jev's paired ratio-of-means
reductions against Sol, including D02--D04 cost reductions of 41.6\%,
42.9\%, and 40.5\%; their intervals appear in Table 7. Controls are
condition-specific, not a pooled Sol baseline. Lower absolute values are
better; axes start at zero with task-specific ranges. Haiku D01 includes
request-limit endpoints.}]{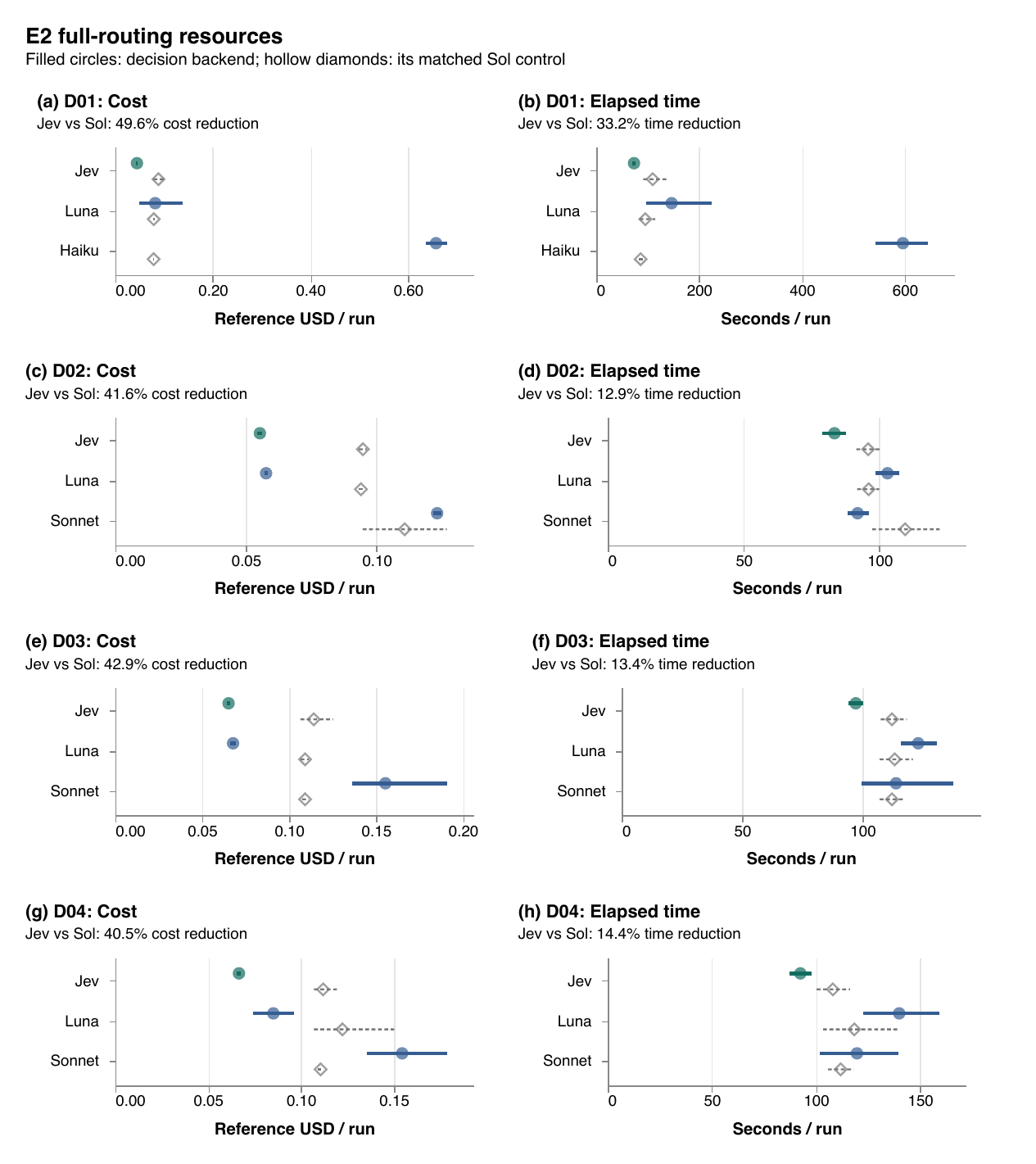}
\caption{E2 comparisons accompanying Table 7. Filled circles show each
decision backend; hollow diamonds show its own matched Sol controls.
Eight subplots retain absolute mean cost (left) and elapsed time
(right), with pointwise 95\% intervals. Subtitles report Jev's paired
ratio-of-means reductions against Sol, including D02--D04 cost
reductions of 41.6\%, 42.9\%, and 40.5\%; their intervals appear in
Table 7. Controls are condition-specific, not a pooled Sol baseline.
Lower absolute values are better; axes start at zero with task-specific
ranges. Haiku D01 includes request-limit endpoints.}
\end{figure}

Jev has the lowest mean cost and elapsed time in all four task-specific
E2 comparisons, including against the inexpensive general-purpose
backends. This repeated pattern, together with all 40 minimum-length Jev
trajectories, supports specialization of the semantic chooser rather
than optimization of Boolean gates alone. The retail results add
evidence in a second application domain and a different action space.

\subsection{Baseline Comparisons and
Ablations}\label{baseline-comparisons-and-ablations}

Haiku and Luna provide lower-cost general-purpose decision baselines,
while Sonnet supplies a further general-purpose comparator in the
expansion. E1 and E2 test two specialization boundaries within the same
Python controller. E1 changes only existence checking; E2 also changes
the content-aware chooser. The E2 gains demonstrate the value of
specializing the richer control problem, rather than relying on a
cheaper tariff alone. Deterministic predicates remain appropriate for
directly observable conditions; semantic routing additionally requires
interpreting requirements, document contents, dependencies, and repair
needs.

\subsection{Statistical
Interpretation}\label{statistical-interpretation}

We report arithmetic means and ratio-of-means reductions, with 100,000
matched-pair bootstrap replicates stratified by BC/CB order. Appendix C
specifies the estimator and seeds. Intervals are exploratory, post hoc,
and pointwise: ten pairs quantify repeated execution within a task, not
ten independent tasks, and do not provide multiplicity-adjusted
superiority tests. Bold entries mark observed extrema. Cap-containing
cells estimate resources to termination. Retail accepted-action counts
are descriptive; the pre-run calibration did not support the planned
noninferiority procedure, as documented in the reproducibility note.

\section{Reproducibility}\label{reproducibility}

We retain immutable configurations, source hashes, initial workspaces,
request and response records, and execution histories. Reference costs
are reconstructed from provider usage. Audits verify model identity,
selected-action dispatch, request counters, and artifact hashes. Paired
runs begin with identical evidence and share policy and candidate
meanings; subsequent generated states may differ. Every reported
observation is traceable to its receipt and terminal outcome.

The accompanying reproducibility note, included with the manuscript
sources, documents cohort construction, collection-level accounting, and
sensitivity analyses. These records are separate from the analyzed-run
resource metric. The manuscript builder checks every reported table
estimate against source analyses, converts citations into
bibliography-linked LaTeX commands, and verifies compilation from a
fresh directory.

Appendix A.2 describes decoder, request-limit, and artifact-integrity
safeguards, together with the network-blocked regression checks.
Synthetic calls are never empirical model-performance observations.

For retail, we retain the case builder, source database and task hashes,
visible inputs, acceptable-action sets, pre-run review, and every
request and response. Both backends receive identical visible decision
content; labels and source identifiers are excluded. The dataset
fingerprint binds the live evaluation to the reviewed 100-case set.
Statistical planning records accompany the descriptive action results.

\section{Discussion}\label{discussion}

\subsection{Specialized Control Improves
Efficiency}\label{specialized-control-improves-efficiency}

The central systems lesson is that cheap decisions and cheap workflows
are different objectives. E1 nearly eliminates the price of a small gate
while leaving the dominant work intact. E2 attacks the content-aware
chooser, but the general-purpose baselines show how repeated generation
can consume the saving. Jev delivers both parts of the improvement:
inexpensive control requests and trajectories that avoid extra
generation in all 40 evaluated E2 runs.

This suggests a concrete design rule: profile the cost of repeated
evidence processing and the expensive actions each decision can
activate, then specialize the high-frequency semantic boundaries.
Preserve deterministic checks for facts that code can establish
directly. Across the four tasks, this division of labor produces
40.5\%--49.6\% lower mean reference cost than matched Sol controls, with
the generation model unchanged. The cost of intelligence is therefore
partly an orchestration problem, not only a model-pricing problem.

\subsection{Input Control and Downstream
Variation}\label{input-control-and-downstream-variation}

We fix initial conditions to evaluate the complete execution induced by
each routing policy. Enforcing identical intermediate artifacts would
suppress downstream differences and change the estimand. Cost and
elapsed-time comparisons are direct system-level measurements that
deliberately include these differences. The retail study additionally
compares decisions on exactly matched visible states.

Even with a common generator, changes in routing and generation can
coexist. A repeat induces additional generation, while model sampling
can change document length and turn count without a routing error. We
therefore report paired outcomes and stage-level measurements without
treating their arithmetic decomposition as causal attribution.

\subsection{Evidence Boundaries}\label{evidence-boundaries}

The software results cover four authored documentation tasks, not
executable software delivery or a representative task population.
Minimum-length completion is an execution property, not proof of optimal
document quality. The retail study demonstrates accepted-action
performance on reviewed same-state cases, but cannot rule out premature
completion in the separate software workflows. Independent blinded
artifact grading and complete customer-resolution evaluation remain
necessary for those claims. Retail labels are project-reviewed and share
source accounts and scenario families, as specified in Section 4.5.

The measurements compare deployed systems. They do not isolate Jev's
architecture, training objective, sampler, tariff, or each individual
routing choice as the causal source of savings. E1 and E2 use fresh
controls and independently generated artifacts; their contrast is a
specialization-boundary comparison, not a fixed-trajectory causal
decomposition. Reference tariffs are not invoices, and client-observed
timings include gateway effects and, in the expansion, shared load. The
statistical scope is stated once in Section 5.9.

\subsection{Beyond Workflow-Step
Routing}\label{beyond-workflow-step-routing}

The reusable problem is broader than selecting the next document step:
map rich evidence and an explicit policy to a bounded action set.
Natural extensions include tool-versus-reply selection, support-queue
assignment and escalation, retrieval-source selection, candidate
filtering and ranking, and semantic validation of prerequisites. SOMA's
reasoning monitor {\citep{ref19}} and Decision4Rec's recommendation
decisions {\citep{ref20}} illustrate adjacent uses of this separation.
The same interface can decide whether to retrieve more evidence, request
a repair, or hand an ambiguous case to a stronger reasoner; permissions
and irreversible actions remain governed by application code.

Stronger decision models could make frequent semantic checks affordable
enough to run throughout an agent's execution, rather than only at
coarse phase boundaries. Better choices could prevent unnecessary
generation, while uncertainty estimates, calibrated on the deployment
task, could reserve expensive reasoning or human review for difficult
states. Independent checks on one state could also be evaluated
together, as supported by Jev's published parallel-output design
{\citep{ref22}.} These are development opportunities, not additional
measured outcomes. The architectural target is clear: specialized
control for frequent bounded judgments, with general-purpose reasoning
and generation where open-ended work is actually needed.

\section{Conclusion}\label{conclusion}

Control-flow routing is not negligible overhead: it repeatedly processes
substantial evidence and determines how often expensive generation is
invoked. Our 480-run workflow study shows why both effects matter.
Existence-only specialization yields limited savings, and low-tariff
general-purpose routers can lose their advantage through longer
trajectories. Jev full routing instead combines a 98.1\% routing-cost
reduction on D01 with minimum-length completion in all 40 E2 runs.
Full-run cost falls by 49.6\% {[}42.9\%, 56.3\%{]} on D01 and
40.5\%--42.9\% on D02--D04. Elapsed time falls by 33.2\% {[}19.6\%,
46.7\%{]} on serial D01 and 12.9\%--14.4\% under the expansion's
concurrent load.

On identical reviewed retail states, Sol and Jev each select accepted
actions in 100/100 cases, while Jev uses 97.88\% less reference cost and
72.30\% less mean request time. Together, these results support
specialized finite-choice control paired with general-purpose generation
across the evaluated software and customer-service settings. The
practical objective is not simply a cheaper router, but a router whose
decisions keep the whole agent efficient.

\section{Data and Code Availability}\label{data-and-code-availability}

Code and data needed to reproduce the reported results are available at
\url{https://github.com/Czk0712/Workflow_routing}. The repository will
be made public upon publication. It contains the experiment code and
tests, frozen task and configuration definitions, run schedules,
sanitized run-level measurements for the 480 workflow experiments and
200 retail requests, and an analysis script that recomputes all reported
point estimates and bootstrap intervals offline. Optional live reruns
are separated from offline reproduction because they require access to
the specified model services and incur provider charges. Credentials,
internal research logs, historical drafts, and raw provider responses
are not distributed.

\bibliography{references}
\clearpage
\appendix

\section{Implementation
Details}\label{appendix-a.-implementation-details}

\subsection{Content-Aware Decision
Rules}\label{a.1-content-aware-decision-rules}

The shared policy applies the following rules to the complete visible
evidence:

\begin{enumerate}
\def\labelenumi{\arabic{enumi}.}
\tightlist
\item
  Inspect actual contents for missing, empty, incorrect, contradictory,
  incomplete, or stale outputs; a filename or a claimed PASS is
  insufficient.
\item
  Repair the earliest unresolved prerequisite using its registered
  producer; select downstream work only when its required inputs are
  usable.
\item
  Repeat the current step for its own defect, or backtrack for an
  upstream defect. Judge downstream staleness from changed content, not
  age alone.
\item
  Select \texttt{complete} only when all current requirements and
  dependencies are satisfied, no mandatory work remains, and the
  execution/existence conditions hold.
\item
  Select \texttt{abstain} for uninterpretable evidence, irreconcilable
  requirements, or no legal repair action; an ordinary missing artifact
  with an available producer is not grounds for abstention.
\end{enumerate}

\subsection{Protocol Integrity and Regression
Checks}\label{a.2-protocol-integrity-and-regression-checks}

Strict decoders validate candidate identity, structured output, model
identity, and usage. The runtime rejects evidence tampering and invalid
completion, and never silently switches the decision backend. Request
counters enforce the same 80-request allowance independently in both
arms. Provider-specific adapters preserve the common choice semantics:
Jev returns native choices and probabilities, while the general-purpose
backends return structured choices. These checks make the measured
terminal outcomes and resource totals reproducible.

We check artifact integrity independently of model decisions. Completed
arms must satisfy the required document paths, ordered headings, line
limits, and initial-file preservation constraints. Our network-blocked
regression tests exercise malformed responses, abstention, premature
termination, backtracking, independent request caps, usage isolation,
and tampered evidence using mocked provider responses. The parallel
extension additionally tests overlapping pairs, isolated receipts,
deterministic collection, and worker-failure draining.

\section{Complete Workflow
Results}\label{appendix-b.-complete-workflow-results}

\subsection{D01 Arm-Level Resources}\label{b.1-d01-arm-level-resources}

Table 8 reports all six D01 conditions, including their separately
collected Sol controls. B denotes the Sol control; C denotes the
comparator assignment in Table 2.

\par\medskip\noindent\begin{minipage}{\linewidth}
\centering
\captionof{table}{D01 arm-level resources (ten valid runs per row). Calls are totals; cost and elapsed time are arithmetic means with pointwise 95\% CIs. Lower resource use is better. Haiku E2 C reaches the request limit in all ten runs; completion is reported separately.}\label{tab:8}
\small\setlength{\tabcolsep}{4pt}\renewcommand{\arraystretch}{1.12}
\sbox{\papertable}{\begin{tabular*}{\linewidth}{@{\extracolsep{\fill}}lrrrrrr@{}}
\toprule
 &  &  & \multicolumn{2}{c}{Reference cost (USD) $\downarrow$} & \multicolumn{2}{c}{Elapsed time (s) $\downarrow$} \\
\cmidrule(lr){4-5}\cmidrule(lr){6-7}
Arm & Completed & Calls & Mean & 95\% CI & Mean & 95\% CI \\
\midrule
\addlinespace[3pt]\multicolumn{7}{l}{\color{PaperBlue}\bfseries E1 Haiku} \\
B & 10/10 & 100 & 0.0792 & \textcolor{PaperMuted}{$[0.0774,\,0.0808]$} & 87.3 & \textcolor{PaperMuted}{$[85.0,\,89.4]$} \\
C & 10/10 & 100 & 0.0759 & \textcolor{PaperMuted}{$[0.0749,\,0.0770]$} & 80.9 & \textcolor{PaperMuted}{$[79.2,\,82.7]$} \\
\addlinespace[3pt]\multicolumn{7}{l}{\color{PaperBlue}\bfseries E1 Jev} \\
B & 10/10 & 100 & 0.0784 & \textcolor{PaperMuted}{$[0.0766,\,0.0801]$} & 85.5 & \textcolor{PaperMuted}{$[82.6,\,88.0]$} \\
C & 10/10 & 100 & 0.0770 & \textcolor{PaperMuted}{$[0.0762,\,0.0778]$} & 82.2 & \textcolor{PaperMuted}{$[80.7,\,83.8]$} \\
\addlinespace[3pt]\multicolumn{7}{l}{\color{PaperBlue}\bfseries E1 Luna} \\
B & 10/10 & 100 & 0.0801 & \textcolor{PaperMuted}{$[0.0789,\,0.0813]$} & 90.8 & \textcolor{PaperMuted}{$[87.5,\,93.9]$} \\
C & 10/10 & 106 & 0.0832 & \textcolor{PaperMuted}{$[0.0757,\,0.0964]$} & 92.1 & \textcolor{PaperMuted}{$[85.8,\,102.1]$} \\
\addlinespace[3pt]\multicolumn{7}{l}{\color{PaperBlue}\bfseries E2 Haiku} \\
B & 10/10 & 100 & 0.0788 & \textcolor{PaperMuted}{$[0.0773,\,0.0803]$} & 86.2 & \textcolor{PaperMuted}{$[82.7,\,90.2]$} \\
C & 0/10 & 800 & 0.6574 & \textcolor{PaperMuted}{$[0.6358,\,0.6797]$} & 596.2 & \textcolor{PaperMuted}{$[542.9,\,645.2]$} \\
\addlinespace[3pt]\multicolumn{7}{l}{\color{PaperBlue}\bfseries E2 Jev} \\
B & 10/10 & 109 & 0.0885 & \textcolor{PaperMuted}{$[0.0784,\,0.1018]$} & 109.7 & \textcolor{PaperMuted}{$[91.2,\,136.7]$} \\
C & 10/10 & 100 & 0.0446 & \textcolor{PaperMuted}{$[0.0434,\,0.0459]$} & 73.3 & \textcolor{PaperMuted}{$[70.6,\,76.0]$} \\
\addlinespace[3pt]\multicolumn{7}{l}{\color{PaperBlue}\bfseries E2 Luna} \\
B & 10/10 & 100 & 0.0791 & \textcolor{PaperMuted}{$[0.0779,\,0.0805]$} & 95.3 & \textcolor{PaperMuted}{$[84.2,\,114.5]$} \\
C & 10/10 & 154 & 0.0820 & \textcolor{PaperMuted}{$[0.0496,\,0.1379]$} & 146.4 & \textcolor{PaperMuted}{$[97.4,\,224.5]$} \\
\bottomrule
\end{tabular*}}
\typeout{PAPER-TABLE-WIDTH:8:\the\wd\papertable:\the\linewidth}
\usebox{\papertable}
\end{minipage}\par\medskip

\subsection{D02--D04 Paired
Contrasts}\label{b.2-d02d04-paired-contrasts}

Table 9 retains every task-condition cell. The comparisons use each
cell's own matched Sol controls; absolute C-arm E2 means appear in Table
7.

\par\medskip\noindent\begin{minipage}{\linewidth}
\centering
\captionof{table}{D02–D04 reductions relative to matched Sol controls. Each cell has ten pairs and 10/10 completed runs per arm. Intervals are pointwise 95\% paired-bootstrap CIs; higher reductions are better, and negative reductions mean increases. Timings reflect concurrent load.}\label{tab:9}
\small\setlength{\tabcolsep}{4pt}\renewcommand{\arraystretch}{1.12}
\sbox{\papertable}{\begin{tabular*}{\linewidth}{@{\extracolsep{\fill}}lrrrr@{}}
\toprule
 & \multicolumn{2}{c}{Full cost reduction (\%) $\uparrow$} & \multicolumn{2}{c}{Elapsed reduction (\%) $\uparrow$} \\
\cmidrule(lr){2-3}\cmidrule(lr){4-5}
Condition & Est. & 95\% CI & Est. & 95\% CI \\
\midrule
\addlinespace[3pt]\multicolumn{5}{l}{\color{PaperBlue}\bfseries D02} \\
E1 Jev & -10.4 & \textcolor{PaperMuted}{$[-44.5,\,15.1]$} & -8.0 & \textcolor{PaperMuted}{$[-30.6,\,9.5]$} \\
E1 Luna & 3.7 & \textcolor{PaperMuted}{$[-12.2,\,17.2]$} & 0.3 & \textcolor{PaperMuted}{$[-12.2,\,10.6]$} \\
E1 Sonnet & -17.6 & \textcolor{PaperMuted}{$[-40.6,\,-1.3]$} & -6.9 & \textcolor{PaperMuted}{$[-23.5,\,5.5]$} \\
E2 Jev & 41.6 & \textcolor{PaperMuted}{$[40.4,\,42.8]$} & 12.9 & \textcolor{PaperMuted}{$[10.9,\,14.9]$} \\
E2 Luna & 38.6 & \textcolor{PaperMuted}{$[37.7,\,39.4]$} & -7.2 & \textcolor{PaperMuted}{$[-13.2,\,-1.1]$} \\
E2 Sonnet & -11.2 & \textcolor{PaperMuted}{$[-31.1,\,3.8]$} & 16.0 & \textcolor{PaperMuted}{$[5.6,\,24.6]$} \\
\addlinespace[3pt]\multicolumn{5}{l}{\color{PaperBlue}\bfseries D03} \\
E1 Jev & 0.7 & \textcolor{PaperMuted}{$[-1.9,\,3.4]$} & 0.6 & \textcolor{PaperMuted}{$[-2.9,\,4.0]$} \\
E1 Luna & 0.4 & \textcolor{PaperMuted}{$[-2.1,\,2.9]$} & -2.5 & \textcolor{PaperMuted}{$[-7.2,\,1.7]$} \\
E1 Sonnet & -2.2 & \textcolor{PaperMuted}{$[-3.9,\,-0.2]$} & 4.9 & \textcolor{PaperMuted}{$[1.7,\,8.2]$} \\
E2 Jev & 42.9 & \textcolor{PaperMuted}{$[38.6,\,47.8]$} & 13.4 & \textcolor{PaperMuted}{$[8.0,\,19.0]$} \\
E2 Luna & 37.9 & \textcolor{PaperMuted}{$[36.4,\,39.6]$} & -8.6 & \textcolor{PaperMuted}{$[-14.3,\,-3.0]$} \\
E2 Sonnet & -42.2 & \textcolor{PaperMuted}{$[-76.4,\,-24.0]$} & -1.5 & \textcolor{PaperMuted}{$[-19.8,\,10.2]$} \\
\addlinespace[3pt]\multicolumn{5}{l}{\color{PaperBlue}\bfseries D04} \\
E1 Jev & 1.1 & \textcolor{PaperMuted}{$[-0.3,\,2.6]$} & -0.6 & \textcolor{PaperMuted}{$[-10.0,\,6.7]$} \\
E1 Luna & -1.0 & \textcolor{PaperMuted}{$[-9.3,\,4.8]$} & -2.3 & \textcolor{PaperMuted}{$[-11.2,\,5.1]$} \\
E1 Sonnet & 0.0 & \textcolor{PaperMuted}{$[-3.9,\,4.6]$} & -5.7 & \textcolor{PaperMuted}{$[-17.0,\,2.8]$} \\
E2 Jev & 40.5 & \textcolor{PaperMuted}{$[37.2,\,45.0]$} & 14.4 & \textcolor{PaperMuted}{$[10.4,\,19.5]$} \\
E2 Luna & 30.4 & \textcolor{PaperMuted}{$[12.5,\,45.6]$} & -18.2 & \textcolor{PaperMuted}{$[-41.9,\,3.3]$} \\
E2 Sonnet & -39.7 & \textcolor{PaperMuted}{$[-60.7,\,-22.8]$} & -7.0 & \textcolor{PaperMuted}{$[-22.1,\,6.5]$} \\
\bottomrule
\end{tabular*}}
\typeout{PAPER-TABLE-WIDTH:9:\the\wd\papertable:\the\linewidth}
\usebox{\papertable}
\end{minipage}\par\medskip

\section{Statistical
Estimation}\label{appendix-c.-statistical-estimation}

For each condition and metric, the reported percentage reduction is the
ratio-of-means contrast

\[R = 100\left(1-\frac{\bar X_C}{\bar X_B}\right).\]

Within each condition, we resample matched pairs with replacement within
BC and CB order strata, preserving both arms. Each of 100,000 replicates
draws five pairs from each stratum. The 2.5th and 97.5th percentiles
define pointwise 95\% intervals. Both cohorts use seed 20261020 plus a
zero-based cell index: D01 sorts condition identifiers, and the
expansion sorts task-condition tuples. Indexing restarts for each
cohort. Table 7 reports C-arm mean intervals. The contrast above is a
ratio of means, not a mean of pairwise percentage changes or a
difference between unmatched runs.

Machine-readable analyses retain full-run and stage-specific costs,
elapsed time, and request counts. Cohort-construction sensitivity
analyses accompany the reproducibility materials; interval
interpretation follows Section 5.9. No resampling or outcome selection
was changed for this presentation.

\end{document}